\pdfoutput=1

\documentclass[11pt]{article}

\usepackage{acl}

\usepackage{times}
\usepackage{latexsym}
\usepackage{tabularx}
\usepackage{fontawesome}
\usepackage[T1]{fontenc}

\usepackage[utf8]{inputenc}

\usepackage{booktabs}

\usepackage{microtype}

\usepackage{inconsolata}
\usepackage{threeparttable}
\usepackage{multirow}
\usepackage{xspace}

\usepackage{graphicx} 

\usepackage{hyperref}
\usepackage{amsmath}

\usepackage{setspace}

\usepackage{makecell}

\usepackage{array}
\usepackage{arydshln}
\usepackage[position=bottom]{subfig}

\usepackage{enumitem}

\newenvironment{itemize*}%
 {\leftmargini=10pt\begin{itemize}%
  \setlength{\itemsep}{0pt}%
  \setlength{\parskip}{0pt}%
  }%
 {\end{itemize}}

\definecolor{weizhey}{rgb}{0.43, 0.71, 0.40}

\usepackage{rotating}
\title{System-Level Natural Language Feedback}

\author{Weizhe Yuan \\
  New York University \\
  \texttt{wy885@nyu.edu} \\\And
  Kyunghyun Cho \\
  New York University \\
  Prescient Design, Genentech \\
  \\\And
  Jason Weston \\
  New York University\\ 
  \\}

\begin{document}
\maketitle
\begin{abstract}
Natural language (NL) feedback offers rich insights into user experience. While existing studies focus on an instance-level approach, where feedback is used to refine specific examples, we introduce a framework for system-level use of NL feedback. We show how to use feedback to formalize system-level design decisions in a human-in-the-loop-process -- in order to produce better models. In particular this is done through: (i) metric design for tasks; and (ii) language model prompt design for refining model responses. We conduct two case studies of this approach for improving search query and dialog response generation, demonstrating the effectiveness of system-level feedback. We show the combination of system-level and instance-level feedback brings further gains, and that human written instance-level feedback results in more grounded refinements than GPT-3.5 written ones, underlying the importance of human feedback for building systems. We release our code and data at \url{https://github.com/yyy-Apple/Sys-NL-Feedback}.
\end{abstract}

\section{Introduction}
Users interacting with a machine learning system offer feedback, either actively or passively. The feedback can be binary ratings \citep{arora2022director}, preference feedback \citep{NEURIPS2020_1f89885d} and natural language (NL) feedback \citep{hancock-etal-2019-learning, scheurer2022training}. Among them, NL feedback is the most general due to its free-form nature, as opposed to the limited choices in other feedback forms. Hence, it is crucial to harness the potential of NL feedback to improve a system.

Existing research on NL feedback typically adopts one of two strategies. The first uses feedback as an auxiliary target in addition to the original task, just like in multitask learning \citep{hancock-etal-2019-learning, https://doi.org/10.48550/arxiv.2208.03270}. The second modifies the original output based on per-instance feedback. The system can either be fine-tuned with the new output \citep{tandon-etal-2022-learning, https://doi.org/10.48550/arxiv.2204.14146} or iteratively self-critique and self-refine at inference time \citep{madaan2023selfrefine, chen2023teaching}. One common limitation of these studies is that they only focus on instance-level learning, where each feedback only serves the instance for which it was received. 
Furthermore, they often assume the availability of feedback for each and every example, which is not practical in real-world scenarios, where feedback is often sparse.

This paper asks the following question: \textit{Can we aggregate instance-level NL feedback to make system-level design decisions that improve language generation systems?} We answer this question by proposing a general framework for aggregating instance-level NL feedback. A set of criteria (i.e., system-level feedback) are first derived from instance-level feedback through a human-in-the-loop process involving clustering and summarization. Those criteria then guide the design of instruction-following language model prompts to refine (i.e., correct) examples, and the development of metrics that align with users' needs. We conduct two case studies of the proposed framework on information-seeking dialog tasks where we improve both the query generator and the response generator of an Internet-augmented dialog system. The experimental results point to the effectiveness of system-level feedback. Our contributions are:
\begin{itemize}
    \item We propose a new method that derives system-level feedback from instance-level feedback, which can guide text generation refinement.
    \item We show how human experts can use system-level feedback to design metrics for evaluating information-seeking dialog systems.
    \item  We demonstrate that combining system-level and instance-level feedback for prompt design yields more helpful refinements for system training w.r.t. the designed metrics above.
   \item 
    We show the importance of human NL feedback by comparing it to GPT-3.5-generated feedback in response refinement. We find that human feedback leads to more grounded refinements that can better guide system learning.
\end{itemize}

\begin{figure*}
    \centering
    \includegraphics[width=0.98\linewidth]{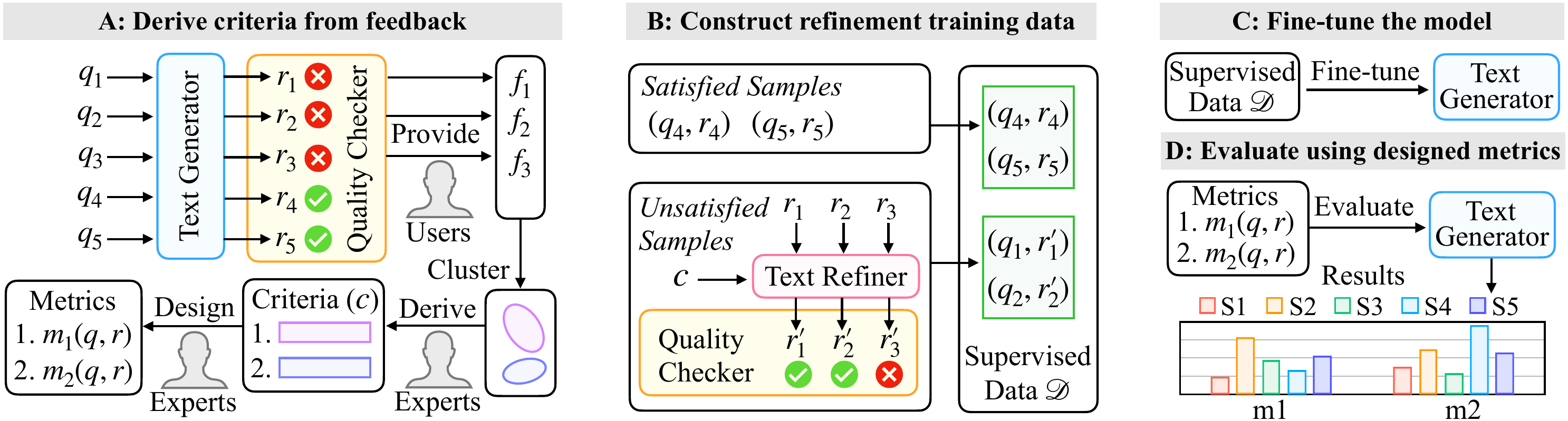}
    \caption{Our framework for incorporating NL feedback
into system-level model design. Using a human-in-the-loop approach, criteria derived from NL feedback guide the creation of prompts for refining responses and metric design to evaluate the improvements.
Notation:
    $q$: query, $r$: response, $f$: feedback, $r^\prime$: refinement, $m(\cdot)$: metric function. $S1\cdots S5$ represent different systems one can compare using this framework. 
    }
    \label{fig:framework}
\end{figure*}

\section{Related Work}
\begin{itemize}[label={}, leftmargin=0pt, labelsep=0pt, itemsep=1pt, labelwidth=0pt]
\item\textbf{Dialog Systems} The rapid development of large language models (LLMs) \citep{DBLP:conf/nips/BrownMRSKDNSSAA20, zhang2022opt} has advanced dialog systems, incorporating techniques like multi-session memory \citep{xu-etal-2022-beyond}, search engine support \citep{komeili-etal-2022-internet}, etc. Recently, ChatGPT's rise has captivated both the NLP community and the public at large. Nowadays, intelligent dialog agents have become an essential part of people's productivity, such as brainstorming \citep{zhang2023visar}, essay polishing \citep{buruk2023academic}, code writing \citep{haensch2023seeing}, etc. However, LLMs also carry potential risks including misinformation \citep{chern2023factool}, sycophancy \citep{sharma2023understanding}, etc., which calls for more thorough evaluations.



\item\textbf{Learning from Human Feedback} As language models increasingly integrate into people's daily life, aligning them with human needs becomes essential \citep{askell2021general}. As a result, researchers have been working on utilizing various human feedback, including preference feedback \citep{NEURIPS2020_1f89885d,ouyang2022training},  binary feedback \citep{li2019don,arora2022director,adolphs2022cringe},  NL feedback \citep{NIPS2016_07563a3f, li2017dialogue, hancock-etal-2019-learning, saunders2022selfcritiquing, scheurer2022training}, and so on. So far, the use of NL feedback is relatively less explored, with most studies focusing on instance-level feedback where each instance receives its own feedback \citep{scheurer2022training, scheurer2023training}. In this work, we propose a general framework for deriving system-level feedback from instance-level feedback, and show the effectiveness of system-level feedback alone and its complementarity with instance-level feedback. 
\end{itemize}



\section{Methodology}
\subsection{Problem Formulation}

Assume we have (1) a text generator $P_{\theta}(r | q)$ that generates a response $r$ to a query $q$, (2) a text refiner $P_{\phi}(r^\prime | r, q, c)$ that generates a refinement $r^\prime$ given the original response $r$, the query $q$, and criteria $c$ that explains what makes a good response, (3) a quality checker $Q(q, r)$ that decides whether $r$ is a satisfactory response given $q$. When deploying $P_{\theta}(r | q)$, for some unsatisfied responses $\mathcal{R}_n=\{r_1, \cdots, r_n\}$, we collect NL feedback for each of them $\mathcal{F}_n=\{f_1, \cdots, f_n\}$. We aim to use $\mathcal{F}_n$ to improve $P_{\theta}(r | q)$ by updating its parameters $\theta$. In our setting, we take the text refiner and quality checker as given. They can either be based on large models like GPT-3 \citep{scheurer2022training} or specialized fine-tuned models \citep{https://doi.org/10.48550/arxiv.2210.15893}.



\subsection{Proposed Framework}
\label{sec:proposed_framework}
Our proposed framework is shown in \autoref{fig:framework}. There are four steps within this framework.

\begin{itemize}[label={}, leftmargin=0pt, labelsep=0pt, itemsep=1pt, labelwidth=0pt]
\item\textbf{Derive criteria from feedback} When deploying the text generator $P_{\theta}(r | q)$, we collect feedback $\mathcal{F}_n$ for some responses $\mathcal{R}_n$. A clustering algorithm is then run (e.g., $k$-means clustering \citep{Hartigan1979}) to identify common issues that can be potentially rectified.
Next, a human-in-the-loop approach is used, where human experts derive a set of criteria $c$ for what constitutes a good response from those clusters. 
These criteria, articulated in natural language, serve as part of the input (prompt) for the text refiner. 
This process relates to prompt engineering in large language models \citep{DBLP:journals/csur/LiuYFJHN23}, where the NL feedback is used to help formalize the prompt engineering process. With these criteria, experts also design metrics \(m_1(\cdot), \cdots, m_k(\cdot)\) to evaluate aspects of user interest.

\item\textbf{Construct refinement training data} 
To improve the text generator, we create a training dataset, \(\mathcal{D}\), that reinforces positive behaviors and rectifies negative ones. If a sample \((q_i, r_i)\) meets \(Q(q_i, r_i) = 1\), it is added to \(\mathcal{D}\) to reinforce good model behavior. Otherwise, the text refiner \(P_{\phi}(r^\prime | r, q, c)\) refines \(r_i\) to \(r_i^\prime\) using prompts based on criteria $c$. If this refined sample \((q_i, r_i^\prime)\) passes \(Q(q_i, r_i^\prime) = 1\), it is added to \(\mathcal{D}\) to modify bad behavior.

\item\textbf{Fine-tune the model} 
After collecting supervised data $\mathcal{D}$, we fine-tune the text generator $P_{\theta}(r | q)$. This data can be combined with existing data that was used to build the baseline deployed system (that did not use feedback).

\item\textbf{Evaluate using designed metrics} 
Finally, we use our designed metrics to assess system performance against user requirements. If successful, the updated system will exhibit improved metrics $m_1(\cdot)$, $\cdots$, $m_k(\cdot)$ compared to the baseline system.
\end{itemize}

\section{Experimental Setup: Dialog Systems}

We study our framework within dialogue system deployment, a context where users naturally offer NL feedback, such as ``that's not correct'' for incorrect responses \citep{https://doi.org/10.48550/arxiv.2210.15893}. Our case studies focus on information-seeking dialogues, where users interact with dialog agents to obtain answers or relevant information \citep{glaese2022improving}.

\begin{itemize}[label={}, leftmargin=0pt, labelsep=0pt, itemsep=1pt, labelwidth=0pt]
\item\textbf{Dialog System Selection} We choose the Blenderbot2 (BB2) dialog system \citep{komeili-etal-2022-internet, xu-etal-2022-beyond} comprised of two modules: 
(1) Query Generator (QG) that generates an Internet search query from dialogue history.
(2) Response Generator (RG) that generates a response using dialogue history and retrieved web documents.\footnote{
We use Google search (\url{https://www.google.com/}) to retrieve the top five relevant documents given a search query.
} We select BB2 because it allows us to study two scenarios: \textit{query generation} and \textit{response generation}.

\item\textbf{Deployment Data} We use the FITS dataset \citep{https://doi.org/10.48550/arxiv.2208.03270} for experiments, which collects diverse feedback from user interactions with Internet-augmented dialogue systems like BB2 and SeeKeR \citep{shuster-etal-2022-language}. Though the dataset includes binary, NL feedback, and gold corrections, we only use binary and NL feedback, given users are less inclined to provide gold corrections for mistakes.


\item\textbf{Text Refiner} Given no gold corrections, we turn to model-based refinement techniques. 
In this work, we use GPT-3.5\footnote{
We use the model \texttt{gpt-3.5-turbo} for our experiments.
} 
as the text refiner and apply greedy decoding during inference.

\item\textbf{Quality Checker}
We train quality checkers for queries and final responses using collected binary feedback. 
Our classifier is based on FLAN-T5\footnote{We use the \texttt{flan-t5-large} model.} \citep{https://doi.org/10.48550/arxiv.2210.11416} trained on 20\% training data, using binary feedback following \citet{https://doi.org/10.48550/arxiv.2210.15893}. We select a threshold to ensure 80\% precision for labels it predicts as positive on the validation set.

\end{itemize}
\begin{table*}[htp]
    \centering 
    \footnotesize
    \renewcommand{\arraystretch}{1}
    \begin{tabular}{cp{0.72\textwidth}cc} 
    \toprule
    \textbf{Group} & \textbf{Feedback type} & \textbf{Num.} & \textbf{\%} \\
    \midrule
    1 & User suggests a search query for Internet search directly. & 2715 & 52.87\% \\
    \midrule
    2 & Suggests specific edits, such as shortening the query or using common words, and so on. & 996 & 19.40\% \\
    \midrule
    3 & Points out that the search query should use keywords instead of copying the original question and should be specific. & 995 & 19.38\% \\
    \midrule
    4 & Points out that the search query is not relevant to the problem. & 429 & 8.35\% \\ 
    \bottomrule
    \end{tabular}
    \caption{
    Case study 1 (query generation): 4 groups of system-level feedback derived from automatic clustering. 
    }
    \label{tab:query_feedback_groups}
\end{table*}

\section{Case Study 1: Query Generation}
\label{sec:scenario1_query_generation}

\begin{table*}[t]
    \centering
    \small
    \setlength\tabcolsep{2.8pt}
    \renewcommand{\arraystretch}{1.2}
    \begin{tabular}{llcccccc}
    \toprule
    \textbf{Type} & \textbf{Criteria (Abbreviated)} & 
    \textbf{NCR} &
    \textbf{Spec.} &
    \textbf{Read.} & 
    \textbf{Con.} &
    \textbf{Cov.} &
    \textbf{Sat.} \\
    \midrule
    (1): Baseline & None & 4.06 & 79.40 & 19.46 & 14.87 & 29.80 & 61.50 \\
    (2): (1)+Rephrase & Rephrase the user's question and keep keywords. & 4.98 & 83.20 & 19.54 & 15.04 & 26.50 & 62.10 \\
    (3): (2)+Specificity & Above + Be accurate and specific for user needs. & 5.00 & \textbf{84.20} & 18.77 & 14.50 & 28.80 & \textbf{63.30} \\
    (4): (3)+Readability & Above + Use simple and common words for better results. & \textbf{5.08} & 80.80 & 19.53 & 15.97 & 29.40 & 62.40 \\
    (5): (4)+Conciseness & Above + Be concise; focus on user's first question. & 4.81 & 80.00 & \textbf{19.70} & \textbf{16.63} & \textbf{35.30} & 62.70 \\
    \bottomrule
    \end{tabular}
    \caption{Case study 1 (query generation): refinement quality via designed metrics when using different criteria to prompt GPT-3.5 for query refinement. Metrics measured: \textbf{NCR}: non-copy rate, \textbf{Spec.}: specificity, \textbf{Read.}: readability, \textbf{Con.}: conciseness, \textbf{Cov.}: coverage. \textbf{Sat.}: satisfaction. The full criteria texts can be found in the Appendix~\ref{app:refinement_with_gpt3.5}.}
    \label{tab:query_instruction_abbr}
\end{table*}

\subsection{Derive Criteria from Feedback} 
\label{sec:query_generation_step1}
We collect all NL feedback from the FITS training split to understand human preferences and derive criteria. We first use SimCSE encoder\footnote{
We use the \texttt{sup-simcse-roberta-large} model.
} 
\citep{gao2021simcse} to encode each feedback. Then, we use $k$-means clustering to group feedback related to query generation into five clusters. From inspecting these (see Appendix~\ref{app:manual_efforts} for detailed manual efforts), we summarize them into four groups (see \autoref{tab:query_feedback_groups}) and derive that a successful search query should (i) rephrase the user's question while keeping important keywords, (ii) be relevant and specific, (iii) use common words for better search coverage, (iv) be concise. The criteria text for crafting the prompt $c$ for the text refiner $P_{\phi}(r^\prime | r, q, c)$ is in \autoref{tab:query_instruction_abbr}.

\subsubsection{Criteria-guided Metric Design}
\label{sec:query-refinement-criteria-guided-metric-design}
Using feedback-derived criteria, we design metrics to mirror users' preferences.\footnote{
When evaluating a set of queries, for a metric defined as a fraction with a constant numerator, we take the average of the denominators of all queries on that metric and take its reciprocal to multiply the numerator.
} Ideally, an effective query should score high across all these metrics.

\begin{itemize}[label={}, leftmargin=0pt, labelsep=0pt, itemsep=1pt, labelwidth=0pt]

\item\textbf{Non-copy rate} measures how much a search query rephrases the user's utterance by examining $n$-gram matching. We define it in \autoref{eq:non-copy-rate} based on BLEU-4 \citep{papineni-etal-2002-bleu} where $s$ is the search query and $u$ is the user question.
\setlength{\abovedisplayskip}{3pt}
\setlength{\belowdisplayskip}{3pt}
\begin{equation}
\label{eq:non-copy-rate}
    \text{Non-copy Rate} = \frac{1}{\text{BLEU-4}(s, u)}
\end{equation}

\item\textbf{Specificity} measures whether the search query sufficiently captures the necessary information to retrieve relevant documents. We use GPT-3.5 as the evaluator \citep{fu2023gptscore}. Details are in the Appendix~\ref{app:evaluation_with_gpt3.5}.

\item\textbf{Readability} measures a search query's clarity based on the word frequency rank (WFR)\footnote{
We use the Kaggle dataset for WFR: \url{https://www.kaggle.com/rtatman/english-word-frequency}
} 
of its terms, as defined in \autoref{eq:readability}, where $w$ is a word in $s$ and $C$ is a scaling constant. Ideally, a query should use common words to improve readability.
\begin{equation}
    \label{eq:readability}
    \text{Readability} = \frac{C}{\text{AVG}_{w \in s}(\text{WFR}(w))}
\end{equation}

\item\textbf{Conciseness} measures the query's brevity by its word count, with its value being the query length's reciprocal, scaled by a constant 100.

\item\textbf{Coverage} measures how specific vs. general a search query is by counting the number of Google search result pages. Considering the wide variation in page count, we employ a relative metric. For refined queries obtained using \autoref{tab:query_instruction_abbr} with the same dialog context, the query with the most results gets a ``Coverage'' score of 1, and others receive 0.

\item\textbf{Satisfaction} measures whether the search query will satisfy the user. It is an overall metric, and we use our trained satisfaction classifier to determine the percentage of satisfied refinements.
\end{itemize}

\subsection{Construct refinement training data}
\label{sec:query_prepare_data}

We sample 1,000 satisfied queries from the FITS training set along with their contexts to add to our supervised training data $\mathcal{D}$. Then, based on \autoref{fig:framework}-(B), for each unsatisfied query $r$, we (1) use GPT-3.5 and criteria $c$ derived from \S\ref{sec:query_generation_step1} to get a refinement $r^\prime$. (2) Use a quality checker to check $r^\prime$'s satisfaction. (3) Add $(q, r^\prime)$ to $\mathcal{D}$ if $r^\prime$ is satisfactory. We elaborate on step (1) in the next section.

\subsubsection{Refinement Generation}
\label{sec:query_refinement_generation}

\begin{table*}[t]
    \centering
    \small
  \setlength\tabcolsep{1.8pt}
  \renewcommand{\arraystretch}{1.2}
  \begin{tabular}{lcccccccccccccccccc}
    \toprule
          & \multicolumn{6}{c}{Valid} & \multicolumn{6}{c}{Test} & \multicolumn{6}{c}{Test Unseen} \\
           \cmidrule(lr){2-7} \cmidrule(lr){8-13} \cmidrule(lr){14-19}
          & 
          \multicolumn{1}{c}{NCR} & \multicolumn{1}{c}{Spec.} & \multicolumn{1}{c}{Read.} & \multicolumn{1}{c}{Con.} & \multicolumn{1}{c}{Cov.} & \multicolumn{1}{c}{Sat.} & \multicolumn{1}{c}{NCR} & \multicolumn{1}{c}{Spec.} & \multicolumn{1}{c}{Read.} & \multicolumn{1}{c}{Con.} & \multicolumn{1}{c}{Cov.} & \multicolumn{1}{c}{Sat.} & \multicolumn{1}{c}{NCR} & \multicolumn{1}{c}{Spec.} & \multicolumn{1}{c}{Read.} & \multicolumn{1}{c}{Con.} & \multicolumn{1}{c}{Cov.} & \multicolumn{1}{c}{Sat.} \\  \midrule

        BB2(QG)  & \textbf{32.8} & 40.5 & \textbf{22.4} & \textbf{32.3} & \textbf{50.6} & 4.8 & \textbf{18.8} & 34.9 & 14.0 & \textbf{34.3} & \textbf{50.9} & 8.8 & \textbf{22.7} & 37.7 & 15.4 & \textbf{32.9} & \textbf{50.3} & 3.2\\
          
          SLT(QG(\faThumbsOUp)) & 2.6 & 60.4 & 19.8 & 21.0 & 30.1 & 9.2 & 2.8 & 58.0 & 17.4 & 22.9 & 30.5 & 12.9 & 3.0 & 55.4 & 18.3 & 22.9 & 31.7 & 7.4\\          
          SLT(QG(\faThumbsOUp+\faThumbsODown)) & 4.8 & \textbf{73.5}  & 22.0 & 18.3 & 19.3 & \textbf{29.6} & 3.8 & \textbf{74.5} & \textbf{21.7} & 18.0 & 18.6 & \textbf{29.0} & 3.6 & \textbf{73.5} & \textbf{19.4} & 17.8 & 18.0 & \textbf{17.2} \\          
        \bottomrule
    \end{tabular}
    \caption{Evaluate query generators on FITS using designed metrics. See \autoref{tab:query_instruction_abbr} caption for abbreviation meanings.
    }
    \label{tab:query_generator_performance_my_metrics}
\end{table*} 
\begin{table}[t]
    \centering
    \small
  \setlength\tabcolsep{3pt}
  \renewcommand{\arraystretch}{1.2}
  \begin{tabular}{@{}lcccccc}
    \toprule
          & \multicolumn{2}{c}{Valid} & \multicolumn{2}{c}{Test} & \multicolumn{2}{c}{Test Unseen} \\ \cmidrule(lr){2-3}\cmidrule(lr){4-5}\cmidrule(lr){6-7}
          & \multicolumn{1}{c}{\textsc{F1}} & \multicolumn{1}{c}{\textsc{PPL}} & \multicolumn{1}{c}{\textsc{F1}} & \multicolumn{1}{c}{\textsc{PPL}} & \multicolumn{1}{c}{\textsc{F1}} & \multicolumn{1}{c}{\textsc{PPL}} \\ \midrule
          BB2(QG) & 9.74 & 16.09 & 14.28 & 9.61 & 16.09 & 10.15 \\
          SLT(QG(\faThumbsOUp)) & 48.63 & 12.83 & 50.51 & 7.64 & 51.75 & 7.84 \\
          SLT(QG(\faThumbsOUp+\faThumbsODown)) & \textbf{51.19} & \textbf{10.34} & \textbf{52.99} & \textbf{7.23} & \textbf{52.21} & \textbf{7.73}\\

          \bottomrule
    \end{tabular}
    \caption{Evaluate query generators on FITS using F1 and perplexity (PPL).
    }
    \label{tab:query_comparison_standard_metrics}
\end{table}
We use GPT-3.5 with criteria-based prompts to refine 1,000 randomly sampled unsatisfied queries (details in Appendix~\ref{app:refinement_with_gpt3.5}). To demonstrate the effectiveness of \autoref{fig:framework}-(A), we conduct ablation studies with different criteria for query refinement. Given our computational budget, for metrics relying on GPT-3.5, we sample 500 dialog contexts and compare the queries resulting from different criteria.

The results are in \autoref{tab:query_instruction_abbr}. Adding criteria in the prompt will shift GPT-3.5's generation, and the performance differences are interpretable using our designed metrics. Specifically, (i) The rephrase criterion increases the non-copy rate. (ii) The relevance criterion increases the relevance metric. (iii) The readability criterion increases the readability and coverage metrics. (iv) Using all the criteria, the refinements achieve reasonably good performance in all our designed perspectives and overall satisfaction. Thus, when collecting training data, we use the four criteria augmented prompt for refinement.




\subsection{Fine-tuning the Model}
\label{sec:query_generator_finetune}
We start from the 400M BB2 query generator and consider two fine-tuning settings: (1) using the satisfied data; and (2) using satisfied and refinement data. During training, we use the Adam optimizer \citep{DBLP:journals/corr/KingmaB14} with a batch size of 8 and learning rate of $7 \times 10^{-6}$ for three epochs. The best checkpoint is chosen based on validation loss.

\begin{table*}[htp]
    \centering 
    \footnotesize
  \renewcommand{\arraystretch}{1}
  \begin{tabularx}{\textwidth}{cp{0.7\textwidth}cc}
    \toprule
    \textbf{Group} & \textbf{Feedback type} & \textbf{Num.} & \textbf{\%} \\
    \midrule
    1 & Clarify his/her demand again. & 3702 & 26.54\%\\
    \midrule
    2 & Complain that the bot (1) does not answer the question or (2) gives irrelevant information or (3) asks the user to find out the answer on his or her own. & 2260 & 16.20\%\\ \midrule
    3 & Point out specific search results that can answer the question. & 2255 & 16.17\%\\ \midrule
    4 & Suggest that the bot should use the search results. & 2130 & 15.27\% \\ \midrule
    5 & States that the answer is (1) factually incorrect, or (2) not grounded on the search results. & 1572 & 11.27\%\\ \midrule
    6& Point out that the bot's answer is not specific/accurate/complete/detailed. & 1309 & 9.39 \%\\ \midrule
    7 & Point out that the bot is not confident in its answers and always begins its responses with ``I am not sure'' or ``I don't know''. & 582 & 4.17\%\\ \midrule
    8 & Complain about repetition/rudeness in bot responses. & 137 & 0.99\%\\

\bottomrule
    \end{tabularx}
    \caption{
    Case study 2 (response generation): 8 groups of system-level feedback derived from automatic clustering. 
    }
    \label{tab:response_feedback_groups}
\end{table*}

\subsection{Evaluation using designed metrics}
We evaluate the following query generators.

\begin{itemize*}
    \item \textbf{BB2(QG)} The original BB2 query generator.
    \item \textbf{SLT(QG(\faThumbsOUp))} System-level trained query generator using only satisfied data.
    \item \textbf{SLT(QG(\faThumbsOUp+\faThumbsODown))} System-level trained query generator using satisfied and refinement data.
\end{itemize*}

\begin{itemize}[label={}, leftmargin=0pt, labelsep=0pt, itemsep=1pt, labelwidth=0pt]
\item\textbf{Results on Standard Metrics} \autoref{tab:query_comparison_standard_metrics} presents the results using standard metrics, as per \citet{https://doi.org/10.48550/arxiv.2210.15893}. Compared to the original BB2 query generator, training with domain-specific data (2\textsuperscript{nd} row) significantly improves F1 word overlap and perplexity metrics. Adding refinement data (3\textsuperscript{rd} row) further enhances these metrics.

\item\textbf{Results on Our Designed Metrics} We also report results on our designed metrics for different query generators in \autoref{tab:query_generator_performance_my_metrics}. It is clear that training on satisfied data produces more specific and satisfactory queries, with further improvements when incorporating refinement data. The original BB2 query generator often generates overly concise queries, hindering the retrieval of the most relevant documents. In other words, although it generates queries that perform well in terms of readability or coverage, it is still an inadequate query generator, as evidenced by the poor satisfaction of the queries it generates. Later, when we refer to ``our trained query generator'', we mean the one trained using both satisfied data and refinement data.
\end{itemize}

\section{Case Study 2: Response Generation}

\begin{table*}[t]
    \centering
    \small
    \setlength\tabcolsep{2pt}
    \renewcommand{\arraystretch}{1.2}
    \begin{tabular}{llcccccc}
    \toprule
    \textbf{Type} & \textbf{Criteria (Abbreviated)} & 
    \textbf{GRD} &
    \textbf{Fact.} &
    \textbf{Help.} & 
    \textbf{Rel.} &
    \textbf{Conf.} &
    \textbf{Sat.} \\
    \midrule
    (1): Baseline & Use a conversational tone; no more than 20 words. & 34.68 & 86.60 & 81.40 & 89.40 & 99.60 & 74.10 \\
    (2): (1)+Groundedness & Above + Use search results to give answers. & 36.81 & 86.60 & 85.00 & 89.00 & \textbf{99.90} & 75.80 \\
    (3): (2)+Relevance & Above + Be concise and targeted, no irrelevant information. & 36.77 & \textbf{88.80} & 85.60 & 89.40 & \textbf{99.90} & 74.90 \\
    (4): (3)+Confidence& Above + Don't start with ``I'm not sure'' or ``I don't know''. & \textbf{39.02} & 87.20 & \textbf{86.60} & \textbf{90.60} & \textbf{99.90} & \textbf{77.00} \\
    \bottomrule
    \end{tabular}
    \caption{Case study 2 (response generation): refinement quality via designed metrics when using different criteria to
prompt GPT-3.5 for response refinement. 
Metrics measured:
\textbf{GRD}: groundedness, \textbf{Fact.}: factuality, \textbf{Help.}: helpfulness, \textbf{Rel.}: relevance, \textbf{Conf.}: confidence. \textbf{Sat.}: satisfaction. The full criteria texts can be found in the Appendix~\ref{app:refinement_with_gpt3.5}.}
    \label{tab:response_instruction_abbr}
\end{table*}

\subsection{Derive criteria from feedback} 

Following the approach in \S\ref{sec:query_generation_step1}, we group all feedback related to response generation into ten clusters. Then, we summarize the following eight groups (see \autoref{tab:response_feedback_groups}) of feedback types by merging some clusters. From \autoref{tab:response_feedback_groups}, we derive that an improved response as indicated by users should (i) ground its answer on relevant search results, (ii) be concise and targeted, (iii) be confident in its answer.
The criteria text for crafting the prompts $c$ for the text refiner 
 $P_{\phi}(r^\prime | r, q, c)$ 
  is given in \autoref{tab:response_instruction_abbr}.

\subsubsection{Criteria-guided Metric Design}
\label{sec:response_generation_criteria_guided_metric_design}
After deriving criteria for response generation from feedback, we design the following metrics to measure the quality of a response as indicated by users.\footnote{
When evaluating a set of responses using one of the following metrics, we take the average of all responses' scores on that metric.
}

\begin{itemize}[label={}, leftmargin=0pt, labelsep=0pt, itemsep=1pt, labelwidth=0pt]

\item\textbf{Groundedness} measures how much the response utilizes the search results by examining $n$-gram matching. We define it in \autoref{eq:groundedness} based on ROUGE-2 \citep{lin-2004-rouge}. Here, $r$ is the response, $d$ is a document from the relevant search set $\mathcal{S}$.
\begin{equation}
\label{eq:groundedness}
    \text{Groundedness} = \max_{d \in \mathcal{S}}\text{ROUGE-2}(r, d)
\end{equation}

\item\textbf{Factuality} checks whether the information in the response is backed by search documents. We use GPT-3.5 with chain-of-thought to measure factuality \citep{luo2023chatgpt}. See Appendix~\ref{app:evaluation_with_gpt3.5} for details.

\item\textbf{Helpfulness} measures whether the response directly answers the user's question. We use GPT-3.5 to measure helpfulness. See Appendix~\ref{app:evaluation_with_gpt3.5} for details.

\item\textbf{Relevance} measures whether the response remains on topic and offers pertinent information. We again use GPT-3.5, with further details in the Appendix~\ref{app:evaluation_with_gpt3.5}.

\item\textbf{Confidence} measures whether the response is in a certain and confident tone. We use simple heuristics to gauge confidence, counting the occurrences of ``I'm not sure'' and ``I don't know.'' If either phrase appears, we consider the response unconfident; otherwise, it's considered confident.

\item\textbf{Satisfaction} measures whether the response satisfies the user, similar to ``satisfaction'' in \S\ref{sec:query-refinement-criteria-guided-metric-design}.

\end{itemize}

\subsection{Construct refinement training data}

\begin{table*}[t]
    \centering
    \small
  \setlength\tabcolsep{1.4pt}
  \renewcommand{\arraystretch}{1.2}
  \begin{tabular}{lcccccccccccccccccc}
    \toprule
          & \multicolumn{6}{c}{Valid} & \multicolumn{6}{c}{Test} & \multicolumn{6}{c}{Test Unseen} \\ 
          \cmidrule(lr){2-7} \cmidrule(lr){8-13} \cmidrule(lr){14-19}
          & 
          \multicolumn{1}{c}{GRD} & \multicolumn{1}{c}{Fact.} & \multicolumn{1}{c}{Help.} & \multicolumn{1}{c}{Rel.} & \multicolumn{1}{c}{Conf.} & \multicolumn{1}{c}{Sat.} & \multicolumn{1}{c}{GRD} & \multicolumn{1}{c}{Fact.} & \multicolumn{1}{c}{Help.} & \multicolumn{1}{c}{Rel.} & \multicolumn{1}{c}{Conf.}  & \multicolumn{1}{c}{Sat.} & \multicolumn{1}{c}{GRD} & \multicolumn{1}{c}{Fact.} & \multicolumn{1}{c}{Help.} & \multicolumn{1}{c}{Rel.} & \multicolumn{1}{c}{Conf.}  & \multicolumn{1}{c}{Sat.} \\  \midrule

          BB2(QG+RG) & 34.1 & 50.0 & 19.0 & 68.2 & 66.8 & 27.1 & 32.4 & 58.3 & 22.0 & 67.8 & 73.7 & 34.9 & 32.9 & 58.4 & 21.8 & 69.0 & 65.7 & 32.1\\
          
          SLT(QG)+BB2(RG) & 39.0 & 66.4 & 26.8 & 74.2 & 80.6 & 33.3 & 35.2 & 58.4 & 29.8 & 71.4 & 83.4 & 40.9 & 37.5 & 59.1 & 30.2& 73.8 & \textbf{77.5} & 37.8\\ 
          
          SLT(QG+RG(\faThumbsOUp)) & 30.6 & 59.1 & 29.2 & 75.6 &76.4  & 35.3 & 27.8 & 53.7 & 31.5 & 69.6 &  80.6 & 41.7 & 29.7 & 60.5 & 31.3 & 73.4 & 72.6  & 39.3\\
          
          SLT(QG+RG(\faThumbsOUp+\faThumbsODown)) & \textbf{48.2} & \textbf{69.1} & \textbf{41.3} & \textbf{81.6} & \textbf{81.1} & \textbf{50.7} & \textbf{43.2} & \textbf{66.7} & \textbf{44.5} & \textbf{76.4} & \textbf{83.6} & \textbf{55.7} & \textbf{45.3} & \textbf{71.6} & \textbf{43.9} & \textbf{79.6} & 76.3 & \textbf{51.4}\\
          \bottomrule
    \end{tabular}
    \caption{Evaluate dialog systems on FITS using designed metrics. See \autoref{tab:response_instruction_abbr} caption for abbreviation meanings.
    }
    \label{tab:performance_my_metrics_system_feedback_only}
\end{table*} 
\begin{table}[t]
    \centering
    \small
  \setlength\tabcolsep{1.9pt}
  \renewcommand{\arraystretch}{1.2}
  \begin{tabular}{@{}lcccccc}
    \toprule
          & \multicolumn{2}{c}{Valid} & \multicolumn{2}{c}{Test} & \multicolumn{2}{c}{Test Unseen} \\ 
          \cmidrule(lr){2-3}\cmidrule(lr){4-5}\cmidrule(lr){6-7}
          & \multicolumn{1}{c}{\textsc{F1}} & \multicolumn{1}{c}{\textsc{PPL}} & \multicolumn{1}{c}{\textsc{F1}} & \multicolumn{1}{c}{\textsc{PPL}}  & \multicolumn{1}{c}{\textsc{F1}} & \multicolumn{1}{c}{\textsc{PPL}} \\ \midrule
          BB2(QG+RG) & 25.78 & 9.40 & 28.30 & 7.41 & 22.99  & 7.75  \\
          SLT(QG)+BB2(RG) & 26.69 & 8.24 & 28.66 & 6.66 & 24.88  & 7.03 \\
          SLT(QG+RG(\faThumbsOUp)) & \textbf{28.20}  & \textbf{7.41} & \textbf{29.73} & \textbf{6.04} & \textbf{25.54} & \textbf{6.43} \\
          SLT(QG+RG(\faThumbsOUp+\faThumbsODown)) & 25.57 & 7.62 & 26.90 & 6.15 & 24.34 & 6.58 \\
          \bottomrule
    \end{tabular}
    \caption{Evaluate dialog systems on FITS via F1 \& PPL.
    }
    \label{tab:performance_standard_metrics_system_feedback_only}
\end{table}
As in \S\ref{sec:query_prepare_data}, we first randomly sample 1,000 satisfied responses together with their contexts to add to our training data $\mathcal{D}$. Then, we go through the following three steps: (1) refinement generation, (2) quality check and (3) collection of filtered data. We will describe (1) in detail in the following section.

\subsubsection{Refinement Generation}
\label{sec:response_refinement_generation}
We use GPT-3.5 with criteria-based prompts to refine 1,000 sampled unsatisfied responses (details in Appendix~\ref{app:refinement_with_gpt3.5}). As in \S\ref{sec:query_refinement_generation}, we conduct ablation studies to demonstrate the effectiveness of derived criteria. The results in \autoref{tab:response_instruction_abbr} highlight: (i) Adding the groundedness criterion improves the groundedness metric. (ii) Adding the relevance criterion increases helpfulness and relevance. (iii) GPT-3.5 refinements are confident and rarely include phrases like ``I'm not sure'' or ``I don't know''. (iv) In terms of satisfaction, the best performance is achieved by the prompt with all criteria added. Therefore, when collecting training data, we use the three criteria-augmented prompt for response refinement.

\subsection{Fine-tuning the Model}
We use the 400M BB2 main model as the baseline response generator and consider two fine-tuning settings:  (1) using only satisfied data; and (2) using both satisfied and refinement data, following
\S\ref{sec:query_generator_finetune}.

\subsection{Evaluation using designed metrics}
\label{subsec:evaluation_using_designed_metrics}
We evaluate the following systems:

\begin{itemize*}
\item \textbf{BB2(QG+RG)} 
Original BB2 response generator paired with the original BB2 query generator.

\item\textbf{SLT(QG)+BB2(RG)} 
Original BB2 response model paired with our system level trained query generator.

\item\textbf{SLT(QG+RG(\faThumbsOUp))} 
Our system-level trained response generator using satisfied data only, paired with our system level trained query generator.

\item\textbf{SLT(QG+RG(\faThumbsOUp+\faThumbsODown))} 
Our system-level trained response generator using satisfied and refinement data, paired with our system level trained query generator.

\end{itemize*}




\begin{table*}[t]
    \centering
    \fontsize{8}{11}\selectfont
  \setlength\tabcolsep{1.5pt}
  \renewcommand{\arraystretch}{1.2}
  \begin{tabular}{lcccccccccccccccccc}
    \toprule
          & \multicolumn{6}{c}{Valid} & \multicolumn{6}{c}{Test} & \multicolumn{6}{c}{Test Unseen} \\ 
          \cmidrule(lr){2-7} \cmidrule(lr){8-13}\cmidrule(lr){14-19}
          & 
          \multicolumn{1}{c}{GRD} & \multicolumn{1}{c}{Fact.} & \multicolumn{1}{c}{Help.} & \multicolumn{1}{c}{Rel.} & \multicolumn{1}{c}{Conf.} & \multicolumn{1}{c}{Sat.} & \multicolumn{1}{c}{GRD} & \multicolumn{1}{c}{Fact.} & \multicolumn{1}{c}{Help.} & \multicolumn{1}{c}{Rel.} & \multicolumn{1}{c}{Conf.} & \multicolumn{1}{c}{Sat.} & \multicolumn{1}{c}{GRD} & \multicolumn{1}{c}{Fact.} & \multicolumn{1}{c}{Help.} & \multicolumn{1}{c}{Rel.} & \multicolumn{1}{c}{Conf.} & \multicolumn{1}{c}{Sat.}\\  \midrule
          
          SLT(QG+RG(\faThumbsOUp+\faThumbsODown)) & 48.2 & \textbf{69.1} & 41.3 & \textbf{81.6} & 81.1 & 50.7 & 43.2 &  66.7 & 44.5 & 76.4 & 83.6 & 55.7 & \textbf{45.3} & 71.6 & 43.9 & 79.6 & 76.3  & 51.4\\
          
          SLT(QG+RG(\faThumbsOUp+HFB\faThumbsODown)) & \textbf{48.8} & 68.1 & \textbf{43.3} & 81.4 & \textbf{91.9} & \textbf{57.3} & \textbf{43.8} & \textbf{68.5} & \textbf{47.8} & \textbf{79.4} & \textbf{93.5} & \textbf{61.2} & 45.0 & \textbf{72.2} & \textbf{45.4} & \textbf{81.2} & \textbf{88.0} & \textbf{57.5}\\
          
          SLT(QG+RG(\faThumbsOUp+GPT3.5FB\faThumbsODown)) & 44.0 & 66.3 & 39.4 & 78.6 & 80.2 & 49.4 & 38.9 & 66.7 & 45.6 & 78.6 & 81.7 & 54.7 & 40.9 & 69.9 & 45.2 & 80.6 & 75.3 & 53.1\\ 
          \bottomrule
    \end{tabular}
    \caption{Case study for combining system-level and instance-level feedback: performance of different dialog systems on FITS datasets, evaluated using our designed metrics. See \autoref{tab:response_instruction_abbr} for the meaning of the abbreviations.
     }
    \label{tab:performance_my_metrics_instance_and_system_level_feedback}
\end{table*}


\begin{itemize}[label={}, leftmargin=0pt, labelsep=0pt, itemsep=1pt, labelwidth=0pt]
\item\textbf{Results on Standard Metrics}
Standard metrics are shown in \autoref{tab:performance_standard_metrics_system_feedback_only}.
Key takeaways include: (i) When using the BB2 response generator, our trained query generator improves the final response quality compared to the BB2 query generator. (ii) Training the response generator on satisfied data leads to further improvements when using our best query generator. (iii) However, training with additional refinement data does not surpass using satisfied data alone. The reason behind (iii) relates to FITS's gold response collection. Often, the gold response is a user-guided, BB2-generated reply. This biases reference-based metrics towards the original BB2 outputs. Moreover, low-quality references may underestimate model performance when using reference-based metrics \citep{zhang2023benchmarking} and we confirmed this with a human evaluation of response quality (see Appendix~\ref{app:human_eval} for details). 


\item\textbf{Results on Our Designed Metrics} \autoref{tab:performance_my_metrics_system_feedback_only} shows the results when using our designed metrics. Notably, (i) when using the BB2 response generator, our trained query generator improves the final response quality from all perspectives compared to the BB2 query generator. (ii) When equipped with our trained query generator, training the response generator on satisfied data leads to consistent improvements in helpfulness compared to the BB2 response generator, indicating the importance of domain-adapted training. (iii) Training the response generator on both satisfied and refinement data improves the final response quality from all perspectives compared to training on satisfied data only, highlighting refinement data's utility in rectifying model errors. (iv) In terms of satisfaction, the best-performing system employs our query and response generators, both trained on satisfied and refinement data. Additionally, as a further baseline, we gathered the first 200 unsatisfied responses into a sparse refinement training set, refined via instance-level feedback. A model trained on this set alongside satisfied data, fell short compared to our system-level trained response generator, as measured by our designed metrics, see Appendix~\ref{app:instance_level_vs_system_level} for details. 



\end{itemize}



\section{Combining System-level Feedback and Instance-level Feedback}

Previous studies \citep{https://doi.org/10.48550/arxiv.2204.14146, https://doi.org/10.48550/arxiv.2210.15893, chen2023improving} have shown the effectiveness of instance-level feedback in the refinement process. To take a step further, we explore the synergy of system-level and instance-level feedback on dialogue systems. Using response generation as a case study, we collect both human and GPT-3.5 feedback (prompt in Appendix~\ref{app:feedback_generation_with_gpt3.5}) for the 1,000 unsatisfied responses from \S\ref{sec:response_refinement_generation}. We then design a refinement prompt integrating both system-level and instance-level feedback, i.e. both the desired criteria and the specific example-based feedback (see Appendix~\ref{app:refinement_with_gpt3.5}). We introduce three systems for comparison.

\begin{itemize*}
    \item\textbf{SLT(QG+RG(\faThumbsOUp+\faThumbsODown))} 
Our system-level trained response generator using satisfied and refinement data, paired with our trained query generator. The system does not use instance-level feedback.

\item\textbf{SLT(QG+RG(\faThumbsOUp+HFB\faThumbsODown))} 
Our system-level trained response generator paired with trained query generator. The response generator is trained on satisfied and refinement data (where we incorporate human-written instance-level feedback (HFB) into the response refinement prompt).

\item\textbf{SLT(QG+RG(\faThumbsOUp+GPT3.5FB\faThumbsODown))} 
Our system-level trained response and query generators, where the response generator is trained on satisfied and refinement data. We incorporate GPT-3.5, rather than human, generated instance-level feedback (GPT3.5FB) into the response refinement prompt.
\end{itemize*}




\subsection{Results of Adding Instance-level Feedback}
Results using our designed metrics are in \autoref{tab:performance_my_metrics_instance_and_system_level_feedback}. We observe that adding human-written feedback to the response refinement part brings improvements in the five criteria-based metrics most of the time, and increases the overall satisfaction consistently. However, adding GPT-3.5 feedback results in degraded performance in groundedness, factuality and confidence. Those observations raise two questions: (1) How does GPT-3.5 feedback differ from human feedback? (2) How does human/GPT-3.5 feedback impact response refinement? We address these questions in subsequent sections.

\subsection{Human vs. GPT-3.5 Feedback Metrics}

To understand why adding human feedback is more beneficial than GPT-3.5 feedback, we analyze their differences through the following perspectives. (1) \textbf{Refinement Success Rate}: Percentage of satisfactory feedback-driven refinements. (2) \textbf{Verbosity}: Average word count of feedback. (3) \textbf{Diversity}: Percentage of unique words. (4) \textbf{Grammar}: Percentage of grammatical feedback sentences.\footnote{
We use Gramformer for grammar error checking: \url{https://github.com/PrithivirajDamodaran/Gramformer}.
}






\begin{figure}[t]
    \centering
    \includegraphics[width=1\linewidth]{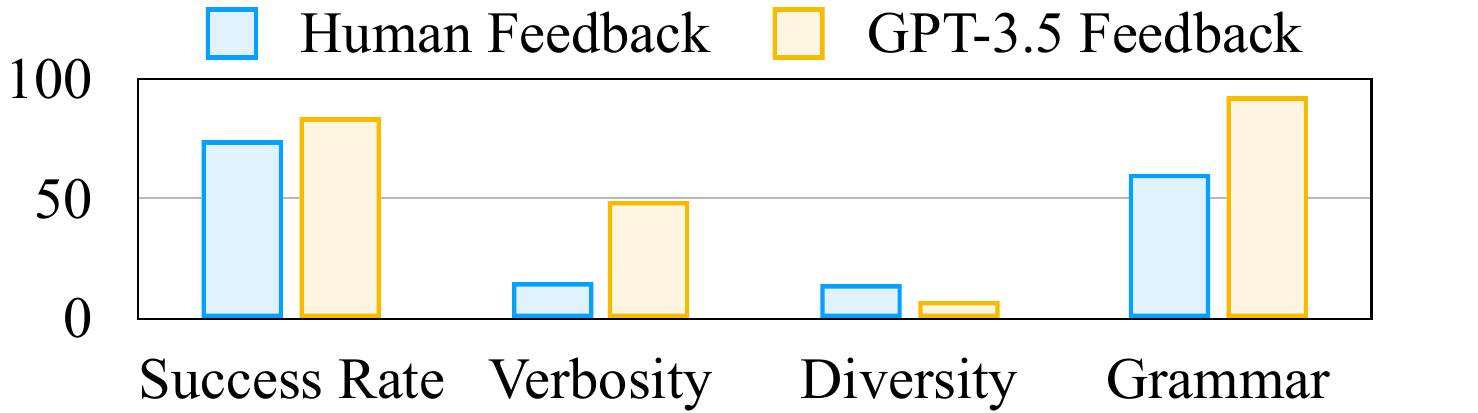}
    \caption{Comparison of human and GPT-3.5 feedback.
    }
    \label{fig:feedback-compare}
\end{figure}
\begin{table}[t]
    \centering
    \small
  \setlength\tabcolsep{3.4pt}
  \renewcommand{\arraystretch}{1.2}
  \begin{tabular}{@{}lcccccc}
    \toprule
       \textbf{Refinement} & \textbf{GRD} & \textbf{Fact.} & \textbf{Help.} & \textbf{Rel.} & \textbf{Conf.} & \textbf{Sat.} \\
        \midrule
           No feedback & 39.16 & \textbf{90.35} & 83.48 & 98.10 & \textbf{100.00} & 76.50 \\ 
           Human FB & \textbf{40.11} & 87.50 & 81.10 & 97.80 & 99.84 & 74.60 \\
           GPT-3.5 FB & 32.77 & 81.50 & \textbf{90.20} & \textbf{98.40} & 99.84 & \textbf{79.50}\\ 
          \bottomrule
    \end{tabular}
    \caption{Quality of refinements with no/human/GPT-3.5 feedback. See \autoref{tab:response_instruction_abbr} for abbreviation meanings.
    }
    \label{tab:compare_instance_feedback_refinements}
\end{table} 


In \autoref{fig:feedback-compare}, we show characteristics of human and GPT-3.5 feedback. Though GPT-3.5 feedback is lengthier and grammatically sound, it lacks the language diversity of human feedback. Upon manual examination, GPT-3.5 feedback is often general, whereas human feedback is direct and specific. See the Appendix~\ref{app:examples_of_gpt3.5_human_instance_level_feedback} for feedback examples.



\subsection{Feedback Impact on Refinements}
\label{sec:feedback_impact_on_refinements}

While GPT-3.5 feedback leads to a higher refinement success rate (see \autoref{fig:feedback-compare}), the performance of the resulting dialog system trained with these refinements falls short w.r.t. all our designed metrics compared to the system trained using human feedback-driven refinements as shown in \autoref{tab:performance_my_metrics_instance_and_system_level_feedback}. Therefore, to understand this further we also evaluate the refinement quality via designed metrics from \S\ref{sec:response_generation_criteria_guided_metric_design}, with results in \autoref{tab:compare_instance_feedback_refinements}. Refinements obtained using human feedback mainly stand out in groundedness and factuality. This aligns with the feedback clusters in \autoref{tab:response_feedback_groups} where over 40\% of the feedback suggests the bot focus more on the search results; that is, focusing more on the search results will make the refinements more grounded, leading to a more grounded final system (see \autoref{tab:performance_my_metrics_instance_and_system_level_feedback}). Since language models are known to hallucinate regardless of their size \citep{DBLP:journals/csur/JiLFYSXIBMF23, li2023halueval}, grounding their generations to the documents is important to ensure factuality. Hence, groundedness of refinements plays an essential role in the performance of trained models.

\subsection{Advantages of Human Feedback}
We find that human feedback pinpoints issues more effectively than GPT-3.5 feedback. For example, when a response does not answer a question, GPT-3.5 will say that the response is unhelpful because it does not contain the information the user wants. In contrast, human feedback often provides specific hints from the search results, guiding the model towards a better response.  Thus, despite GPT-3.5 producing seemingly informative feedback, it currently can't match the nuance of human annotators.

\section{Conclusion}
In this paper, we present a framework that harnesses system-level NL feedback. By using a set of instance-level feedback, we derive system-level feedback for refinement prompt engineering and metric design. We show the effectiveness of system-level feedback through two case studies: generating queries and formulating dialogue responses. We further combine system-level and instance-level feedback in the refinement data construction process, and observe that the resulting trained response generator makes considerable improvements versus either alone. Finally, we explore the possibility of substituting instance-level human feedback with GPT-3.5 feedback. We find that human feedback stands out in capturing main issues, while GPT-3.5 feedback is lengthy and less focused. 


\section{Limitations}
Due to the lack of publicly available natural language feedback datasets, our experiments were limited to the small-scale dialog system BB2, which does not represent the current state-of-the-art. We recognize that integrating more advanced models such as ChatGPT could yield further insights, presenting a promising direction for future research. As relevant datasets become more accessible, we look forward to exploring these possibilities.

\section{Acknowledgement}
The work was done as part of the Meta–NYU mentorship program and partly supported by the National Science Foundation (under NSF Award 1922658). Kyunghyun Cho is supported by the Samsung Advanced Institute of Technology (under the project Next Generation Deep Learning: From Pattern Recognition to AI).




\bibliography{anthology,custom}

\appendix

\section{Appendix}
\label{sec:appendix}

\subsection{Manual Efforts Required to Derive System-level Criteria}
\label{app:manual_efforts}
In our approach, the feedback grouping was a semi-automated process. Initially, we employed k-means clustering, utilizing the SimCSE encoder to categorize the feedback sentences. This clustering process was conducted once without human intervention. Then we employed a streamlined, non-iterative manual approach for cluster curation. Specifically, two domain experts independently reviewed 50 samples (each requires around 30s to read) from each of the 15 clusters (including clusters for queries and responses), requiring approximately 375 minutes per person for this phase. This was followed by a collaborative discussion to merge insights and remove duplicate clusters, amounting to an additional 60 minutes per expert. Thus, the total human effort amounted to approximately 14.5 person-hours.

\subsection{Refinement with GPT-3.5}
\label{app:refinement_with_gpt3.5}
We instruct GPT-3.5 to generate query refinements and response refinements using carefully crafted prompts, as shown in \autoref{tab:query_refinement_prompt} and \autoref{tab:response_refinement_prompt}. The corresponding ablation studies with full criteria text for query refinement and response refinement are shown in \autoref{tab:query_instruction_full} and \autoref{tab:response_instruction_full}. The prompt for using both system-level feedback and instance-level feedback for response refinement is shown in \autoref{tab:response_refinement_prompt_with_instance_feedback}.

\begin{table*}[t]
    \centering
    \footnotesize
  \renewcommand{\arraystretch}{1}
  \begin{tabularx}{\textwidth}{@{}X@{}}
    \toprule
    \textbf{Prompt for query refinement with GPT-3.5} \\ \midrule
    Given the dialog history, your task is to refine the original search query used to search the Internet so that the modified search query will search for documents that better match the user's needs. \underline{You should follow the following requirements:} \\\textbf{\texttt{[Criteria]}}\\
    Below is the dialog context. \\\textbf{\texttt{[Dialog Context]}}\\
    Below is the bot's unsatisfactory query.\\
    \textbf{\texttt{[Original Query]}}\\
    You should modify the original search query into the following: 
     \\
\bottomrule
    \end{tabularx}
    \caption{Case study 1: prompt for query refinement with GPT-3.5. \textbf{\texttt{[Criteria]}}, \textbf{\texttt{[Dialog Context]}} and \textbf{\texttt{[Original Query]}} are placeholders to be filled. The underlined sentence is removed when \textbf{\texttt{[Criteria]}} is None.
    }
    \label{tab:query_refinement_prompt}
\end{table*}

\begin{table*}[t]
    \centering
    \footnotesize
  \begin{tabularx}{\textwidth}{@{}X@{}}
    \toprule
    \textbf{Prompt for response refinement with GPT-3.5} \\ \midrule
    Given the dialog history and the unsatisfactory last response the bot gave, your task is to modify the response appropriately to keep the conversation fluent and consistent. You should follow the following requirements: \\\textbf{\texttt{[Criteria]}}\\
    Below is the dialog context. \\\textbf{\texttt{[Dialog Context]}}\\
    Below is the bot's unsatisfactory response.\\
    \textbf{\texttt{[Original Response]}}\\
    Below are some useful search results that you could use.\\
    \textbf{\texttt{[Search Documents]}}\\
    You should modify the original response into the following: 
     \\
\bottomrule
    \end{tabularx}
    \caption{Case study 2: prompt for response refinement with GPT-3.5. \textbf{\texttt{[Criteria]}}, \textbf{\texttt{[Dialog Context]}}, \textbf{\texttt{[Original Response]}} and \textbf{\texttt{[Search Documents]}} are placeholders to be filled.
    }
    \label{tab:response_refinement_prompt}
\end{table*}

\begin{table*}[t]
    \centering
  \setlength\tabcolsep{1pt}
  \renewcommand{\arraystretch}{1}
  {\fontsize{8}{10}\selectfont
  \begin{tabular}{p{0.1\textwidth}p{0.55\textwidth}p{0.052\textwidth}p{0.052\textwidth}p{0.052\textwidth}p{0.052\textwidth}p{0.052\textwidth}p{0.052\textwidth}}
    \textbf{Type} & \textbf{Criteria} & \begin{turn}{61}\textbf{Non-copy}\end{turn} & \begin{turn}{61}\textbf{Specificity}\end{turn} & \begin{turn}{61}\textbf{Readability}\end{turn} & \begin{turn}{61}\textbf{Conciseness}\end{turn} & \begin{turn}{61}\textbf{Coverage}\end{turn} & \begin{turn}{61}\textbf{Satisfaction}\end{turn} \\
    \midrule
Baseline & None & 4.06 & 79.40 & 19.46 & 14.87 & 29.80 & 61.50 \\
\midrule
Baseline +Rephrase & (1) To better adapt to search engines, it is best not to copy the user's original words directly. You can rephrase the user's question, use some keywords for the search, and if the user mentions some abbreviations, restore them to their full names. & 4.98 & 83.20  & 19.54 & 15.04 & 26.50 & 62.10 \\
\midrule
Baseline  +Rephrase  +Specificity & (1) To better adapt to search engines, it is best not to copy the user's original words directly. You can rephrase the user's question, use some keywords for the search, and if the user mentions some abbreviations, restore them to their full names. (2)  Be accurate and specific enough to reflect the user's needs. &  5.00 & \textbf{84.20} & 18.77 & 14.50 & 28.80 & \textbf{63.30} \\
\midrule
Baseline  +Rephrase +Specificity  +Readability  & (1) To better adapt to search engines, it is best not to copy the user's original words directly. You can rephrase the user's question, use some keywords for the search, and if the user mentions some abbreviations, restore them to their full names. (2)  Be accurate and specific enough to reflect the user's needs. (3) To be able to search for more results, you should use more simple and commonly used words. & \textbf{5.08} & 80.80  & 19.53 & 15.97 & 29.40 & 62.40 \\
\midrule
Baseline +Rephrase +Specificity +Readability +Conciseness & (1) To better adapt to search engines, it is best not to copy the user's original words directly. You can rephrase the user's question, use some keywords for the search, and if the user mentions some abbreviations, restore them to their full names. (2)  Be accurate and specific enough to reflect the user's needs. (3) To be able to search for more results, you should use more simple and commonly used words. (4) Your search query should be concise. If the user asks multiple questions, you should focus on his/her first question. & 4.81 & 80.00 & \textbf{19.70} & \textbf{16.63} & \textbf{35.30} & 62.70 \\
\bottomrule
    \end{tabular}  }
    \caption{Case study 1 (query generation): refinement quality via designed metrics when using different criteria to prompt GPT-3.5 for query refinement.
    }
    \label{tab:query_instruction_full}
\end{table*}

\begin{table*}[t]
    \centering
    \footnotesize
  \setlength\tabcolsep{1pt}
  \renewcommand{\arraystretch}{0.1}
    {\fontsize{8}{10}\selectfont
  \begin{tabular}{p{0.12\textwidth}p{0.55\textwidth}p{0.05\textwidth}p{0.05\textwidth}p{0.05\textwidth}p{0.05\textwidth}p{0.05\textwidth}p{0.05\textwidth}}
    \textbf{Type} & \textbf{Criteria} & 
    \begin{turn}{65}\textbf{Groundedness}\end{turn} & 
    \begin{turn}{65}\textbf{Factuality}\end{turn} & 
    \begin{turn}{65}\textbf{Helpfulness}\end{turn} & \begin{turn}{65}\textbf{Relevance}\end{turn} & \begin{turn}{65}\textbf{Confidence}\end{turn} & \begin{turn}{65}\textbf{Satisfaction}\end{turn}\\
    \midrule
Baseline & (1) The modified response should be conversational in tone and no more than twenty words. & 34.68 & 86.60 & 81.40 & 89.40 & 99.60 & 74.10\\

\midrule
Baseline \newline+Groundedness & (1) The modified response should be conversational in tone and no more than twenty words. (2) If the user asks a question, you should use relevant search results to answer the user’s question correctly. Please do not let the user check out some resources on his or her own. &  36.81 & 86.60 & 85.00 & 89.00 & \textbf{99.90} & 75.80 \\

\midrule
Baseline  \newline+Groundedness  +Relevance & (1) The modified response should be conversational in tone and no more than twenty words. (2) If the user asks a question, you should use relevant search results to answer the user’s question correctly. Please do not let the user check out some resources on his or her own. (3) Your modified response should be as concise and targeted as possible, and not include additional information the user has not asked for. & 36.77 & \textbf{88.80} & 85.60  & 
 89.40 & \textbf{99.90} & 74.90 \\

\midrule
Baseline  \newline+Groundedness +Relevance  +Confidence  & (1) The modified response should be conversational in tone and no more than twenty words. (2) If the user asks a question, you should use relevant search results to answer the user’s question correctly. Please do not let the user check out some resources on his or her own. (3) Your modified response should be as concise and targeted as possible, and not include additional information the user has not asked for. (4) Please be confident in your response, and don’t start your response with “I’m not sure” or “I don’t know”. & \textbf{39.02} & 87.20 & \textbf{86.60}  & \textbf{90.60} & \textbf{99.90} & \textbf{77.00}\\

\bottomrule
    \end{tabular}}
    \caption{
    Case study 2 (response generation): refinement quality via designed metrics when using different criteria to
prompt GPT-3.5 for response refinement.
    }
    \label{tab:response_instruction_full}
\end{table*}
\subsection{Evaluation with GPT-3.5}
\label{app:evaluation_with_gpt3.5}
Since previous studies have demonstrated GPT-3's capability in the evaluation of aspects such as factuality \citep{luo2023chatgpt}, helpfulness \citep{fu2023gptscore}, relevance \citep{fu2023gptscore}, etc. We use GPT-3.5 to evaluate the following perspectives with the help of \texttt{ChatEval} \citep{Yuan_chateval_2023}.
\paragraph{Query Specificity}
\begin{table*}[t]
    \centering
    \footnotesize
  \begin{tabularx}{\textwidth}{@{}X@{}}
    \toprule
    \textbf{Prompt for response refinement with GPT-3.5 (with instance-level feedback)} \\ \midrule
    Given the dialog history and the unsatisfactory last response the bot gave, your task is to modify the response appropriately to keep the conversation fluent and consistent. You should follow the following requirements:\\\textbf{\texttt{[Criteria]}}\\
    Below is the dialog context. \\\textbf{\texttt{[Dialog Context]}}\\
    Below is the bot's unsatisfactory response.\\
    \textbf{\texttt{[Original Response]}}\\
    Below is the feedback for the bot's unsatisfactory response.\\
    \textbf{\texttt{[Feedback]}}\\
    Below are some useful search results that you could use.\\
    \textbf{\texttt{[Search Documents]}}\\
    You should modify the original response into the following: 
     \\
\bottomrule
    \end{tabularx}
    \caption{Prompt for response refinement with GPT-3.5 (with instance-level feedback). \textbf{\texttt{[Criteria]}}, \textbf{\texttt{[Dialog Context]}}, \textbf{\texttt{[Original Response]}}, \textbf{\texttt{[Feedback]}} and \textbf{\texttt{[Search Documents]}} are placeholders to be filled.
    }
    \label{tab:response_refinement_prompt_with_instance_feedback}
\end{table*}
\begin{table*}[t]
    \centering
    \footnotesize
  \begin{tabularx}{\textwidth}{@{}X@{}}
    \toprule
    \textbf{Prompt for query specificity evaluation with GPT-3.5} \\ \midrule
      You are evaluating a search query for a dialog using a specific set of standards. Below is the dialog context.\\
      \texttt{\textbf{[Dialog Context]}} \\
      Below is the search query.\\
      \texttt{\textbf{[Query]}}\\
      Below are the criteria.\\
      Decide whether the search query is accurate and specific enough to enable retrieval of the most relevant documents on the Internet that are sufficient to answer the user's question.\\
      \\
    Does the search query meet the criterion? First, write out in a step-by-step manner your reasoning about the criterion to be sure that your conclusion is correct. Avoid simply stating the correct answers at the outset. Then print only the single character "Y" or "N" (without quotes or punctuation) on its own line corresponding to the correct answer. At the end, repeat just the letter again by itself on a new line.\\
  Reasoning:
     \\
\bottomrule
    \end{tabularx}
    \caption{Prompt used to let GPT-3.5 evaluate query specificity. \textbf{\texttt{[Dialog Context]}} and \textbf{\texttt{[Query]}} are placeholders to be filled.
    }
    \label{tab:query_specificity_gpt3.5_prompt}
\end{table*}

We use GPT-3.5 to measure specificity \citep{fu2023gptscore} where we concatenate the dialog context and search query together and ask GPT-3.5 to judge whether the search query is specific using the chain-of-thought technique \citep{NEURIPS2022_9d560961}. 
In particular, we use the prompt as shown in \autoref{tab:query_specificity_gpt3.5_prompt}. Before applying it to measure the quality of query refinements. We manually labeled 50 search queries from the FITS training split, and each one's specificity label was decided by three annotators through majority vote. We calculate the agreement between GPT-3.5 and human annotators, and the result is 80\%.

\paragraph{Response Factuality}
\begin{table*}[t]
    \centering
    \footnotesize
  \begin{tabularx}{\textwidth}{@{}X@{}}
    \toprule
    \textbf{Prompt for response factuality evaluation with GPT-3.5} \\ \midrule
      You are evaluating a response for a dialog using a specific set of standards. Below is the dialog context.\\
      \texttt{\textbf{[Dialog Context]}} \\
      Below are some search documents that may help continue this dialog.\\
      \texttt{\textbf{[Search Documents]}}\\
      Below is the response.\\
      \texttt{\textbf{[Response]}}\\
      Below is the criteria.\\
      Determine if the information in the response can be found in one or more search documents.\\
      \\
    Does the response meet the criterion? First, write out in a step-by-step manner your reasoning about the criterion to be sure that your conclusion is correct. Avoid simply stating the correct answers at the outset. Then print only the single character "Y" or "N" (without quotes or punctuation) on its own line corresponding to the correct answer. In the end, repeat just the letter again by itself on a new line.\\
  Reasoning:
     \\
\bottomrule
    \end{tabularx}
    \caption{Prompt used to let GPT-3.5 evaluate response factuality. \textbf{\texttt{[Dialog Context]}}, \textbf{\texttt{[Search Documents]}} and \textbf{\texttt{[Reponse]}} are placeholders to be filled.
    }
    \label{tab:response_factuality_gpt3.5_prompt}
\end{table*}
We concatenate the search documents and response, and ask GPT-3.5 to judge if all information in the response is supported by the search documents. The prompt we use is shown in \autoref{tab:response_factuality_gpt3.5_prompt}. We conducted a meta-evaluation where we asked three NLP PhD students at the same university as the first author to manually label 50 responses from the FITS training split. The annotation guideline we showed them is the same as the prompt designed for GPT-3.5 evaluation. Then, the three annotators decided on each one's factuality label through a majority vote. The agreement between GPT-3.5 and human annotators is 88\%.

\paragraph{Response Helpfulness}
\begin{table*}[t]
    \centering
    \footnotesize
  \begin{tabularx}{\textwidth}{@{}X@{}}
    \toprule
    \textbf{Prompt for response helpfulness evaluation with GPT-3.5} \\ \midrule
      You are evaluating a response for a dialog using a specific set of standards. Below is the dialog context.\\
      \texttt{\textbf{[Dialog Context]}} \\
      Below is the response.\\
      \texttt{\textbf{[Response]}}\\
      Below are the criteria.\\
      Does the answer directly solve the question?\\
      \\
    Does the response meet the criterion? First, write out in a step-by-step manner your reasoning about the criterion to be sure that your conclusion is correct. Avoid simply stating the correct answers at the outset. Then print only the single character "Y" or "N" (without quotes or punctuation) on its own line corresponding to the correct answer. In the end, repeat just the letter again by itself on a new line.\\
  Reasoning:
     \\
\bottomrule
    \end{tabularx}
    \caption{Prompt used to let GPT-3.5 evaluate response helpfulness. \textbf{\texttt{[Dialog Context]}} and \textbf{\texttt{[Reponse]}} are placeholders to be filled.
    }
    \label{tab:response_helpfulness_gpt3.5_prompt}
\end{table*}
The prompt we use is shown in \autoref{tab:response_helpfulness_gpt3.5_prompt}. We conducted a meta-evaluation where we manually labeled 30 responses from the FITS training split and 20 responses from the Red Team dataset \citep{bai2022training}. Three annotators decided on each response's helpfulness label through majority vote. The agreement between GPT-3.5 and human annotators is 84\%.

\paragraph{Response Relevance}
\begin{table*}[t]
    \centering
    \footnotesize
  \begin{tabularx}{\textwidth}{@{}X@{}}
    \toprule
    \textbf{Prompt for response relevance evaluation with GPT-3.5} \\ \midrule
      You are evaluating a response for a dialog using a specific set of standards. Below is the dialog context.\\
      \texttt{\textbf{[Dialog Context]}} \\
      Below is the response.\\
      \texttt{\textbf{[Response]}}\\
      Below are the criteria.\\
      Is the response relevant to the topic at hand? It's essential to recognize that the response does not need to be highly specific to the preceding question. As long as it remains focused on the topic at hand, it is considered relevant.\\
      \\
    Does the response meet the criterion? First, write out in a step-by-step manner your reasoning about the criterion to be sure that your conclusion is correct. Avoid simply stating the correct answers at the outset. Then print only the single character "Y" or "N" (without quotes or punctuation) on its own line corresponding to the correct answer. In the end, repeat just the letter again by itself on a new line.\\
  Reasoning:
     \\
\bottomrule
    \end{tabularx}
    \caption{Prompt used to let GPT-3.5 evaluate response relevance. \textbf{\texttt{[Dialog Context]}} and \textbf{\texttt{[Response]}} are placeholders to be filled.
    }
    \label{tab:response_relevance_gpt3.5_prompt}
\end{table*}
 The prompt we use is shown in \autoref{tab:response_relevance_gpt3.5_prompt}. We conducted a meta-evaluation where we manually labeled 30 responses from the FITS training split and 20 responses from the Red Team dataset \citep{bai2022training}. Three annotators decided on each response's relevance label through majority vote. The agreement between GPT-3.5 and human annotators is 84\%.


\subsection{Human Evaluation on Response Outputs}
\label{app:human_eval}

In \S\ref{subsec:evaluation_using_designed_metrics}, our analysis revealed that training the response generator with both satisfied data and refinement data does not yield superior performance over using satisfied data alone, as evidenced by the F1 score and Perplexity (PPL) metrics. We hypothesized that this outcome might be attributed to a data bias in the FITS dataset, wherein the gold standard references are frequently produced by the BB2 model. Consequently, standard reference-based metrics, such as F1, tend to favor responses that closely resemble BB2 outputs. This bias potentially results in the underestimation of performance for models generating responses deviating from the BB2 distribution. 

\begin{figure}[t]
    \centering
    \includegraphics[width=1\linewidth]{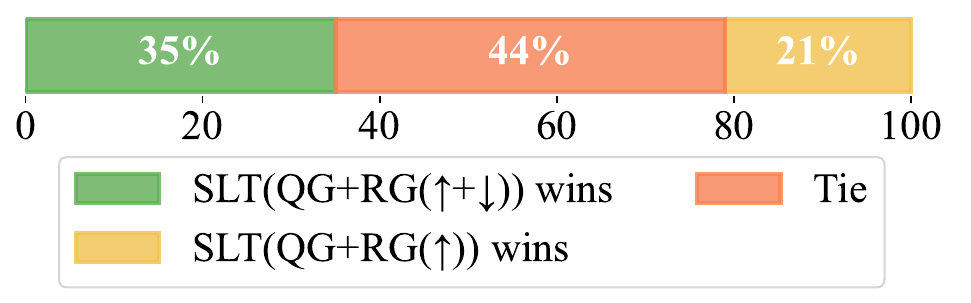}
    \caption{Win rates for system trained with both satisfied and refinement data and system trained with satisfied data only.
    }
    \label{fig:win-rate}
\end{figure}

To address this limitation, we expanded our evaluation methodology beyond model-based metrics. We conducted an additional human evaluation to compare 100 responses generated by SLT(QG+RG(\faThumbsOUp)) and SLT(QG+RG(\faThumbsOUp+\faThumbsODown)) against the same queries. In this evaluation, two human annotators were asked to select their preferred response from the two provided, with the options including a ``tie''. If both annotators agreed that one response was superior, the corresponding model was awarded a ``win''. In cases of disagreement or agreement on ties, the outcome was recorded as a tie. The results of this human evaluation, presented in \autoref{fig:win-rate}, indicate that SLT(QG+RG(\faThumbsOUp+\faThumbsODown)) achieved a higher win rate compared to SLT(QG+RG(\faThumbsOUp)). This finding confirms our hypothesis that reference-based metrics alone are insufficient for evaluating this task, highlighting the need for more robust metrics in system assessment.

\subsection{Instance-level Feedback vs. System-level Feedback}
\label{app:instance_level_vs_system_level}
We argue that one of the drawbacks of instance-level approaches that utilize NL feedback is that they typically assume that every instance receives a feedback text, which is not practical in the real world where feedback tends to be sparse. Therefore, we also conducted a comparison experiment that assumes sparse instance-level feedback. Specifically, we collected the first 200 unsatisfied responses into a sparse refinement training set, refined via instance-level feedback only. We then train the response generator on this set alongside the satisfied data and compare its performance to our system-level trained model. \autoref{tab:instance-vs-system-level-fb} shows the performance of the two models as measured by our designed metrics. The system-level trained response generator outperforms the sparse instance-level trained response generator by a large margin on all metrics, demonstrating the importance of system-level feedback in a sparse instance-level feedback setting.

\begin{table*}[t]
    \centering
    \small
  \setlength\tabcolsep{1.4pt}
  \renewcommand{\arraystretch}{0.8}
  \begin{tabular}{lcccccccccccccccccc}
    \toprule
          & \multicolumn{6}{c}{Valid} & \multicolumn{6}{c}{Test} & \multicolumn{6}{c}{Test Unseen} \\ 
          \cmidrule(lr){2-7} \cmidrule(lr){8-13} \cmidrule(lr){14-19}
          & 
          \multicolumn{1}{c}{GRD} & \multicolumn{1}{c}{Fact.} & \multicolumn{1}{c}{Help.} & \multicolumn{1}{c}{Rel.} & \multicolumn{1}{c}{Conf.} & \multicolumn{1}{c}{Sat.} & \multicolumn{1}{c}{GRD} & \multicolumn{1}{c}{Fact.} & \multicolumn{1}{c}{Help.} & \multicolumn{1}{c}{Rel.} & \multicolumn{1}{c}{Conf.}  & \multicolumn{1}{c}{Sat.} & \multicolumn{1}{c}{GRD} & \multicolumn{1}{c}{Fact.} & \multicolumn{1}{c}{Help.} & \multicolumn{1}{c}{Rel.} & \multicolumn{1}{c}{Conf.}  & \multicolumn{1}{c}{Sat.} \\  \midrule
          
          SLT(QG+RG(\faThumbsOUp+\faThumbsODown)) & \textbf{48.2} & \textbf{69.1} & \textbf{41.3} & \textbf{81.6} & \textbf{81.1} & \textbf{50.7} & \textbf{43.2} & \textbf{66.7} & \textbf{44.5} & \textbf{76.4} & \textbf{83.6} & \textbf{55.7} & \textbf{45.3} & \textbf{71.6} & \textbf{43.9} & \textbf{79.6} & \textbf{76.3} & \textbf{51.4}\\
        ILT(QG+RG(\faThumbsOUp+\faThumbsODown)) & 35.5 & 59.9 & 30.2 & 76.2 & 69.1 & 38.2 & 32.4  &  53.2 & 36.2 & 73.2 & 74.4 & 44.5 & 34.7 & 57.9 & 35.0 & 76.2 &  66.6 & 42.4 \\
          \bottomrule
    \end{tabular}
    \caption{Evaluate dialog systems on FITS using designed metrics. \textbf{SLT(QG+RG(\faThumbsOUp+\faThumbsODown))}: Our system-level trained response generator using satisfied and refinement data, paired with our trained query generator. \textbf{ILT(QG+RG(\faThumbsOUp+\faThumbsODown))}: Our instance-level trained response generator using satisfied and sparse refinement data, paired with our trained query generator.
    }
    \label{tab:instance-vs-system-level-fb}
\end{table*}

\subsection{Feedback Generation with GPT-3.5}
\label{app:feedback_generation_with_gpt3.5}
\begin{table*}[t]
    \centering
    \footnotesize
  \begin{tabularx}{\textwidth}{@{}X@{}}
    \toprule
    \textbf{Prompt for feedback generation with GPT-3.5} \\ \midrule
    Given the dialog history, the unsatisfactory last response the bot gave, and the requirements for a good response, your task is to write detailed and constructive feedback to improve the unsatisfactory response. The requirements for a good response include the following: \\
    (1) The response should be conversational in tone and no more than twenty words. \\
    (2) If the user asks a question, the response should use relevant search results to answer the user’s question correctly. It should not leave the user to check out some resources on his or her own. \\
    (3) The response should be as concise and targeted as possible, and not include additional information the user has not asked for. \\
    (4) The response should be in a confident tone and not start with “I’m not sure” or “I don’t know”.\\
\\
    Below is the dialog context. \\\textbf{\texttt{[Dialog Context]}}\\
    Below is the bot's unsatisfactory response.\\
    \textbf{\texttt{[Original Response]}}\\
    The feedback for the unsatisfactory response is the following:
     \\
\bottomrule
    \end{tabularx}
    \caption{Prompt for feedback generation with GPT-3.5. \textbf{\texttt{[Dialog Context]}} and \textbf{\texttt{[Original Response]}} are placeholders to be filled. 
    }
    \label{tab:feedback_generation_prompt}
\end{table*}
Previous studies have demonstrated the capability of large language models to generate informative and useful feedback \citep{madaan2023selfrefine, chen2023teaching}. Therefore, we also investigate using GPT-3.5 to generate instance-level feedback for each unsatisfied response. The prompt we use for feedback generation is in \autoref{tab:feedback_generation_prompt}.

\subsection{Examples of GPT-3.5 and Human Instance-level Feedback}
\label{app:examples_of_gpt3.5_human_instance_level_feedback}
We list examples of instance-level feedback written by humans and GPT-3.5 in \autoref{tab:human_gpt3.5_instance_level_feedback}.

\begin{table*}[htp]
    \centering 
    \footnotesize
  \setlength\tabcolsep{8pt}
  \renewcommand{\arraystretch}{0.9}
  \begin{tabularx}{\textwidth}{XX}
    \toprule
    \textbf{Human written instance-level feedback} & \textbf{GPT-3.5 written instance-level feedback} \\
    \midrule
      Good. You just need to include the reason and his age at the time of his death in addition to this for a complete answer. &     The response is not relevant to the user's question and includes additional information that the user did not ask for. A good response should focus on answering the user's question about how Fitzgerald died, using relevant search results. It should be concise, confident, and conversational in tone.\\
      \midrule
      You could tell me to take my cat to the vet when they get sick. &    The response is not helpful. Please provide information on treatment options for cat rhinovirus. Use relevant search results to answer the user's question correctly and be concise in your response. Avoid providing additional information the user has not asked for.\\
      \midrule
      The bot did not list Azerbaijan or Estonia, which were included in the search results. The bot should include these two other Soviet Republics in the response. &  The response is informative but could be improved. Instead of listing the republics in a long sentence, try breaking them down into bullet points for easier reading. Also, it would be helpful to mention that there were 15 republics in total, as the user requested.\\
      \midrule
      Again, the bot's "answer" doesn't really make sense. It claims that a character, not an actor, played the part of Celie. &    The response is confusing and irrelevant. Please provide a clear and concise answer to the user's question using relevant search results. Avoid adding unnecessary information or asking unrelated questions.\\
      \midrule
      The answer is right there in the results, and Charles de Gaulle isn't correct. &      The response is good, but it could be improved by being more conversational and concise. Instead of just stating the answer, try something like "Charles de Gaulle was the first French president in 1848. Is there anything else I can help you with?" This makes the response more engaging and shows confidence in the answer.\\
      \midrule
      Interesting. We need to know which premier specifically, with data to back up the claim. Can we be more specific? &      The response is not helpful as it does not answer the user's question. Please provide a specific answer with relevant search results to support it. Avoid guessing or providing irrelevant information. Be confident and concise in your response.\\
      \midrule
      Search about living creatures specifically. &     The response is not relevant to the user's question. Please provide a direct answer to the user's question using relevant search results. Avoid providing additional information that the user has not asked for. Be confident in your response and avoid starting with "I'm not sure" or "I don't know".\\
      \midrule
      Actually it looks like that happened way later. Let's go back to the Revolutionary War and how that affected Britain. &     The response is not helpful as it does not answer the user's question. A better response would be to use relevant search results to explain the reasons for the decline of the British empire, such as economic challenges, decolonization movements, and global power shifts. The response should be confident and concise, and not include additional information the user has not asked for.\\
      \midrule
      Use the search results provided to formulate some answer. Mention how there are wave patterns that one meteorologist uses and got 2/3 correct, mention a tool that they use like the doppler radar which looks for things like airborne tornado debris. &     The response is too general and does not answer the user's question. Instead, use relevant search results to provide specific weather conditions that meteorologists look for to predict tornado formation. Keep the response concise and confident in tone.\\
      \midrule
      One of the search results specifies that Paul Tillich was Rollo May's mentor and friend, confirming that they did know each other. &    The response should have used relevant search results to answer the user's question. It should have been more targeted and concise, and not included additional information the user did not ask for. A better response would have been: "I'm not sure if they knew each other, but according to my search results, Tillich was a major influence on Rollo May's work."\\
      \midrule
      Why didn't you mention CrossFit or Tabata? &    The response should have used relevant search results to answer the user's question correctly. Instead of saying "I'm not sure," the bot should have confidently provided the correct answer. Additionally, the response should have been more concise and targeted, only providing the specific exercise type that burns the most calories.\\

\bottomrule
    \end{tabularx}
    \caption{
    Examples of human written and GPT-3.5 written instance-level feedback. For each row, both human feedback and GPT-3.5 feedback are written to the same unsatisfied response.
    }
    \label{tab:human_gpt3.5_instance_level_feedback}
\end{table*}

\end{document}